\documentclass[conference]{IEEEtran}

\IEEEoverridecommandlockouts

\usepackage{cite}
\usepackage{amsmath,amssymb,amsfonts}
\usepackage{amsfonts}
\usepackage{algorithmic}
\usepackage{mathtools}
\usepackage{graphicx}
\usepackage{textcomp}
\usepackage{xcolor}

\usepackage{subcaption}
\usepackage{multirow}

\newcommand{\emb}{\mathbf}

\newcommand{\entity}[1]{\mbox{\textsf{#1}}}

\def\BibTeX{{\rm B\kern-.05em{\sc i\kern-.025em b}\kern-.08em
T\kern-.1667em\lower.7ex\hbox{E}\kern-.125emX}}

\newcommand{\R}{\mathbb{R}}

\newcommand{\comment}[1]{}

\begin{document}

\title{Clustering and Network Analysis for the Embedding Spaces of Sentences and Sub-Sentences}

\author{\IEEEauthorblockN{Yuan An}
\IEEEauthorblockA{
\textit{Metadata Research Center} \\
\textit{College of Computing and Informatics} \\
\textit{Drexel University, Philadelphia, PA} \\
ya45@drexel.edu}
\and
\IEEEauthorblockN{Alexander Kalinowski}
\IEEEauthorblockA{
\textit{Metadata Research Center} \\
\textit{College of Computing and Informatics} \\
\textit{Drexel University, Philadelphia, PA} \\
ajk437@drexel.edu}
\and
\IEEEauthorblockN{Jane Greenberg}
\IEEEauthorblockA{
\textit{Metadata Research Center} \\
\textit{College of Computing and Informatics} \\
\textit{Drexel University, Philadelphia, PA}\\
jg3243@drexel.edu}
}

\maketitle

\begin{abstract}
Sentence embedding methods offer a powerful approach for working with short
textual constructs or sequences of words. By representing sentences 
as dense numerical vectors, many natural language processing (NLP) 
applications have improved their performance. 
However, relatively little is understood about the latent structure of sentence 
embeddings. Specifically, research has not addressed whether the length and structure of  
sentences impact the sentence embedding space and topology.  
This paper reports research on a set of comprehensive clustering and network analyses 
targeting sentence and sub-sentence embedding spaces.
Results show that one method generates the most clusterable embeddings. 
In general, the embeddings of span sub-sentences have better clustering properties 
than the original sentences. The results have implications 
for future sentence embedding models and applications.

\end{abstract}
    

%
%
\section{Introduction}
\label{sec:introdcution}

The good properties of word embeddings \cite{word2vec,glove} have inspired the development of 
various methods \cite{sentencebert, gem, infersent, skipthought, quickthought, short-text-embedding} 
for sentence embeddings which represent short text or
sentences, i.e., sequences of words and symbols, as dense numerical vectors.
Many downstream natural language
processing (NLP) applications such as semantic textual similarity (STS) \cite{sentencebert},
sentiment analysis \cite{evaluation-sentence-embeddings}, service recommendation \cite{improved-weighted-removal},
and relation extraction \cite{sentence-embedding-alignment}
have utilized sentence embeddings for improving
their performance. 
However, relatively little is understood about the latent structure of 
sentence embeddings. There are two main reasons. First, sentences have variable 
lengths, composed of innumerable combinations of individual words with 
ambiguous boundaries. Second, there are a variety of sentence embedding methods 
guided by different principles.
In particular, sentence embedding methods range 
from simple word embedding aggregation to sophisticated deep encoder-decoder 
neural networks \cite{infersent}. More recent techniques include 
transformer-based BERT-like models \cite{sentencebert}. Consequently, 
the resultant sentence embedding vectors encode different types of information for different purposes.

\begin{figure*}[!th]
	\centering
	\subfloat[Embeddings of the Original Sentences]{
		\includegraphics[width=0.5\linewidth]{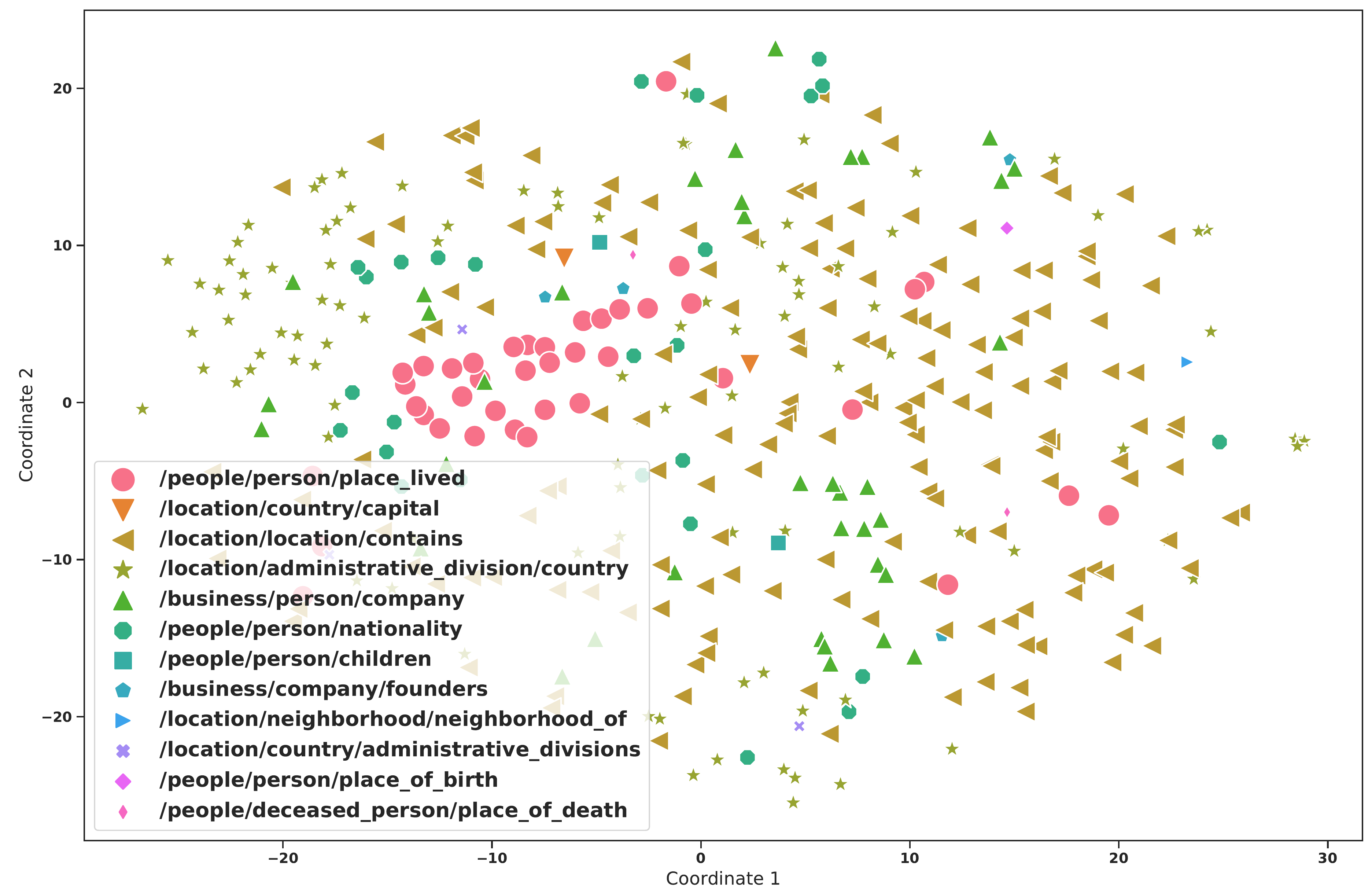} 
		\label{fig:sBERT-sentence}
	}
	\subfloat[Embeddings of the Spanning Sub-Sentences]{
		\includegraphics[width=0.5\linewidth]{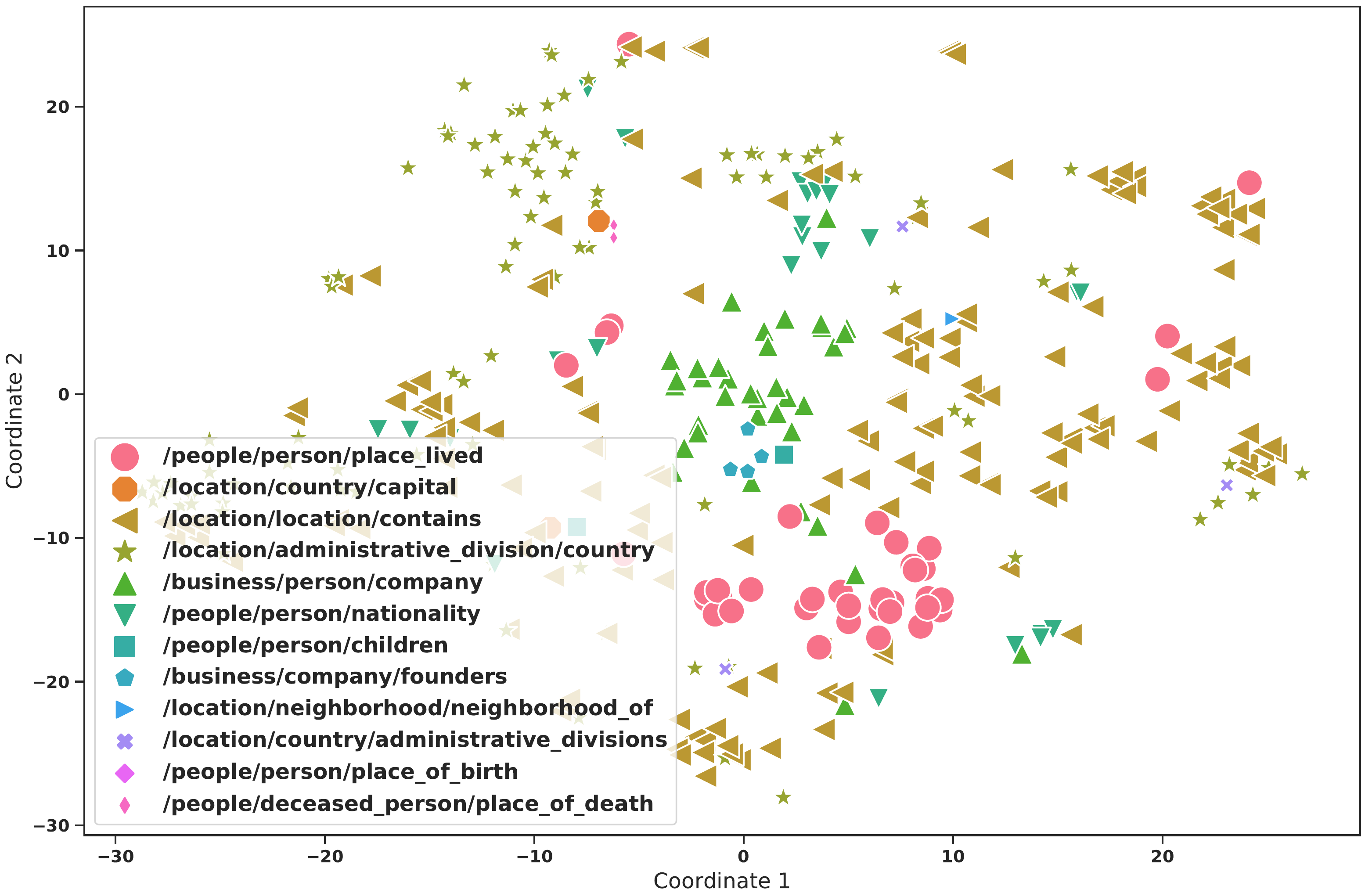}
		\label{fig:sBERT-span}
	}
	\caption{A 2-D visualization of the embeddings generated by SentenceBERT on the test set of the NYT dataset.
		The shapes represent the embeddings of sentences in (a) and sub-sentences in (b). Different shapes 
		correspond to different relations used to label the sentences and sub-sentences. 
	A cluster of the same shapes indicates the embeddings of the sentences and sub-sentences expressing the same relation 
	are located close to each other. The clearer boundaries between the clusters of shapes in (b) indicate a better clusterability
	than the embeddings in (a) in terms of relation labels. 
	}
	\label{fig:sBERT-sentence-span}
\end{figure*}   

Past effort has been made to explore properties of sentence embeddings. 
The work in \cite{fine-grained-sentence-embeddings} 
evaluates the predictive ability of sentence embeddings through predictive tasks. 
Another work in \cite{sentence-analogies} performs a sentence
analogy task by evaluating the degree to which lexical
analogies are reflected in sentence embeddings. 
A significant issue in the current studies is 
that the sentences are mostly simple sentences.
\emph{It remains to be answered whether lengths and structures of sentences have any impacts on
the topology of sentence embedding spaces.} 
  
In this paper, we explore the latent structure of the embedding spaces of complex sentences and their sub-sentences
with regard to capturing semantic regularities. 
We leverage the widely-used dataset for the study of relation extraction \cite{RiedelYM10,cotype,cnn-pcnn-att}.
The dataset was constructed by distantly aligning relations
in Freebase \cite{freebase} with sentences from the New York
Times (NYT) corpus of years 2005-2006. Each sentence in the dataset 
is labeled by a semantic relation between two entity mentions in the
sentence. The data set is 
well-suited for our study because we can use the entity mentions as
anchors to study the embeddings of various sub-sentences.
We collect a set of representative sentence embedding methods and apply these methods to 
the original NYT sentences and various sub-sentences containing the entity mentions. 
Our goal is to examine whether the embeddings of (sub-)sentences expressing 
the same semantic relation cluster together. 
If the clustering signal in the latent structure is strong,
unsupervised methods can recognize relations by simply measuring
the distances between embedding vectors, which will greatly reduce the bottleneck of collecting labeled
training data. 
  
The work we present in this paper shows some methods generate more clusterable embeddings than others.
The sentence structures also have effects on the clusterability of resultant embeddings. 
As an illustration, Figure \ref{fig:sBERT-sentence-span} shows a 2-D visualization of 
the embedding spaces generated by the SentenceBERT method on 
the test set of the NYT dataset. The embedding space of the span sub-sentences in  
Figure \ref{fig:sBERT-sentence-span}(b) has better clustering structures than 
that of the original sentences in 
Figure \ref{fig:sBERT-sentence-span}(a). 
For reproducibility, all the data and code of the study in this paper are put in a public repository\footnote{\label{repository}https://github.com/sent-subsent-embs/clustering-network-analysis}.

The rest of the paper is structured as follows. Section \ref{sec:related-work} discusses
related work. 
Section \ref{sec:background} describes the technical background of the sentence embedding methods. 
Section \ref{sec:sentence-sub-sentence} presents the methods for extracting sub-sentences.
Section \ref{sec:analytic-design} describes the analytic methods and experimental process.
Section \ref{sec:experimental-results} presents the experimental results.
Section \ref{sec:discussion} discusses the study.
Finally, Section \ref{sec:conclusion} concludes the paper.

%
%
\section{Related Work}
\label{sec:related-work}

There is an extensive body of research on exploring properties of word embeddings   
\cite{levy-goldberg-2014-linguistic,word2vec,graph-based-exploration}. 
The work in \cite{graph-based-exploration} applied graph theoretic tools on
the semantic networks induced by word embeddings. The study demonstrated that network analysis
on word embeddings could draw interesting structures about semantic similarity between words. 
In contrast, exploring the properties of sentence embeddings 
has been much more limited. A common approach is to compare the 
performance of sentence embeddings in various downstream NLP tasks as in \cite{fine-grained-sentence-embeddings}. 
The first task of SemEval 2014 \cite{semeval2014, semeval2014-svm}
applied and evaluated sentence embeddings in a semantic relatedness task. 
The authors in \cite{white-how-well2015} evaluated sentence 
embeddings using a semantic classification task. The RepEval 2017 Shared Task 
\cite{repEval2017-multi-genre, repEval2017-LCT}
compared seven sentence embedding methods on shared sentence entailment task.
The work in \cite{evaluation-sentence-embeddings}
performed a comprehensive evaluation of sentence embedding methods using a wide variety of
downstream and linguistic feature probing tasks.
Another work \cite{comparative-sentence-embedding-paraphrasing} 
evaluated several sentence embedding methods, including BERT, InferSent, 
semantic nets and corpus statistics
(SNCS), and Skipthought, by performing a paraphrasing task. 
All the work attempted to discover the relationships between the geometric structures of 
embedding spaces and NLP applications. 

Our work is closely related to the study of sentence analogy in \cite{sentence-analogies}.
Differing from the work, our work evaluates
and explores the clustering properies of sentence embeddings with
regard to expressing semantic relations between entities.
More importantly, our work explores the latent structures of the embeddings of complex sentences and their sub-sentences.

%
%
%
\section{Sentence Embedding Methods}
\label{sec:background}

In this study, we select a set of representative methods that can be classified into 3 categories: 
\emph{word-embedding-aggregation} approach, \emph{from-scratch-sentence-embedding} approach, and 
\emph{pre-trained-fine-tune} approach. 
The classification is based on the guiding principles and algorithms employed by
the methods. Here, we briefly describes the technical details of the methods in each category.
The public repository contains
all implementation code and pointers to the original sources.

\noindent 
\textbf{Category 1: word-embedding-aggregation}:\\
\emph{Method 1. GloVe-Mean \cite{fasttext}}: 
GloVe-Mean takes the arithmetic average of the word embeddings in a sentence as  
the sentence embedding. As its name indicates, GloVe-Mean uses the pre-trained
word embeddings generated by the GloVe method \cite{glove}. 
Let $s=\{w_{1}, w_{2}, ..., w_{n}\}$ be a sentence consisting of a sequence of words,
$w_{1}$, $w_{2}$, ..., $w_{n}$. Let $\emb{v}(w_{i})\in \R^d$ be the $d$-dimensional embedding vector of a word $w_{i}$. 
The GloVe-Mean generates
the embedding vector $\emb{v}(s)\in \R^d$ for the sentence $s$ as:
\[
\emb{v}(s) =  \frac{{\sum}_{i=1}^{n} \emb{v}(w_{i})}{n}
\]
GloVe-Mean is one of the simplest ways to convert the embeddings of a sequence of words
to a single sentence embedding. 
In our study, we use our own home-grown code for GloVe-Mean.

\noindent
\emph{Method 2. GloVe-DCT \cite{discretecosine}}: GloVe-Mean treats the words in a sentence as in a bag, ignoring their ordering.  
To address this, GloVe-DCT stacks individual GloVe word vectors $\emb{v}(w_{1})$, $\emb{v}(w_{2})$,... , 
$\emb{v}(w_{n})$ into a $n\times d$ matrix. It then applies a discrete cosine transformation (DCT) on the columns.
Given a vector of real numbers $c_0, \ldots , c_N$, DCT calculates a sequence of coefficients as follows:
\begin{center}
	$$coef[0] = \sqrt{\frac{1}{N}}\sum_{n=0}^N c_n$$
\end{center}
and
\begin{center}
	$$coef[k] = \sqrt{\frac{2}{N}}\sum_{n=0}^N c_n \cos \frac{\pi}{N} (n + \frac{1}{2})k$$
\end{center}
The choice of $k$ typically ranges from 1 to 6, where a choice of zero is essentially similar to GloVe-Mean. 
To get a fixed-length and consistent 
sentence vector, GloVe-DCT extracts and concatenates the first $K$ DCT coefficients and discards higher-order 
coefficients. The size of sentence embeddings is $Kd$.
In our study, we use our own home-grown code for GloVe-DCT.

\noindent
\emph{Method 3. GloVe-GEM \cite{gem}}: 
For a sentence 
$s=\{w_{1}, w_{2}, ..., w_{n}\}$, 
GloVe-GEM takes the word embeddings $\{\emb{v}(w_{i})\in \R^d, i = 1..n\}$ as input. 
It generates the sentence embedding $\emb{v}(s)$ by 
a weighted sum
\[
\emb{v}(s) = \sum_{i=1}^{n} \alpha_{i} \emb{v}(w_{i}) \quad\mathrm{s.t.}\quad \alpha_{i} = \alpha_{n}+\alpha_{s}+\alpha_{u}
\]
where the weights $\alpha_{i}, i=1..n$ come from three scores: a novelty score $\alpha_{n}$, a
significance score $\alpha_{s}$, and a corpus-wise uniqueness score $\alpha_{u}$. 
To compute the three scores, 
GloVe-GEM applies the Gram-Schmidt Process (also known as QR factorization) 
to the context matrices of the words in the sentence.
For each word, its context matrix is made up with the word embeddings in its surrounding context.
The method then builds 
an orthogonal basis of the context matrix. The scores of the word 
are computed based on principled measures using the bases.
Finally, GloVe-GEM removes the sentence-dependent principle components from the weighted sum. 
In our study, we use our own home-grown code for GloVe-GEM too.

\noindent
\textbf{Category 2: from-scratch-sentence-embedding}: \\
\emph{Method 4. Skipthought \cite{skipthought}}: Inspired by the skip-gram model of word2vec, Skipthought generates
sentence embeddings via the task of predicting neighboring sentences. Skipthought 
depends on a training corpus of contiguous text. It thus uses a large collection
of novels, the BookCorpus unlabeled dataset, for training its model. The model is 
in the encoder-decoder framework. An encoder maps
words to a sentence vector and a decoder is used to generate the surrounding sentences. 
Several choices of encoder-decoder pairs have been explored, including ConvNet-RNN, RNN-RNN,
and LSTM-LSTM. 
In our study, we use the open source implementation of Skipthought\footnote{https://github.com/ryankiros/skip-thoughts}.

\noindent
\emph{Method 5. Quickthought \cite{quickthought}}: Quickthought also uses the unlabeled BookCorpus for training its 
model. Instead of reconstructing the surface form of 
the input sentence or its neighbors, Quickthought uses the embedding of the current sentence to
predict the embeddings of neighboring sentences. In particular, given a sentence, Quickthought's 
model chooses the correct target sentence from a set of candidate sentences.
The model achieves this by replacing the generative
objectives with a discriminative approximation.
In our study, we use the open source implementation of 
Quickthought\footnote{https://github.com/lajanugen/S2V}.

\noindent
\emph{Method 6. InferSentV1 and Method 7. InferSentV2 \cite{infersent}}: InferSent 
uses a three-way classifier to predict the degree of
sentence similarity (similar, not similar, neutral). 
It builds a bi-directional LSTM model
pre-trained on natural language inference (NLI) tasks.
InferSent comes in two flavors, a
V1 model using the pre-trained GloVe vectors and a V2 model
using the pre-trained FastText \cite{fasttext} vectors. Individually, 
we refer to them as \emph{InferSentV1}
and \emph{InferSentV2}.
In our study, we use the open source implementation of 
InferSent\footnote{https://github.com/facebookresearch/InferSent}.

\noindent
\emph{Method 8. LASER \cite{laser}}: LASER trains a Bi-directional LSTM
model on a massive scale, multilingual corpus. 
It uses parallel sentences accross 93
input languages. LASER is able to focus on mapping
semantically similar sentences to close areas of the embedding
space. It allows the model to focus more on meaning and less
on syntactic features. Each layer of the LASER model is 512
dimensional. By concatenating both the forward and backward representations,
LASER generates a final sentence embedding of dimension 1024. 
In our study, we use the open source implementation of 
LASER\footnote{https://github.com/facebookresearch/LASER}.

\noindent
\textbf{Category 3: pre-trained-fine-tune}:\\
\emph{Method 9. SentenceBERT \cite{sentencebert}}: 
BERT \cite{bert} like pre-trained language models have helped 
many NLP tasks achieve state-of-the-art results. One issue of BERT is that it does not
directly generate sentence embeddings.  SentenceBERT \cite{sentencebert}
is a modification of the pre-trained
BERT network. It uses siamese and triplet network
structures to derive semantically meaningful
sentence embeddings. Specifically, SentenceBERT derives a fixed sized sentence embedding
by adding a pooling operation to the output
of BERT / RoBERTa. The network structure depends on the available
training data. A variety of structures and objective functions are tested, including
{\emph{Classification Objective Function}, 
{\emph{Regression Objective Function},
and {\emph{Triplet Objective Function}.
In our study, we use the open source implementation of 
SentenceBERT\footnote{https://github.com/UKPLab/sentence-transformers}.
In particular, we use the base model 
``bert-base-nli-mean-tokens"\footnote{https://huggingface.co/sentence-transformers/bert-base-nli-mean-tokens}.  
The model computes the mean of all output vectors of the BERT.

%
%
\section{Sentence and Sub-Sentence}
\label{sec:sentence-sub-sentence}

Sentence segmentation \cite{effective-semi-supervised-sentence-segmentation}
is a non-trivial NLP task that aims to divide text into meaningful component sentences.
Automatic sentence segmentation typically divides text
based on syntactic structures such as punctuation. 
The resultant sentences often express multiple 
ideas with variable lengths. 
In this study, we examine the strengths and weaknesses of 
different methods for encoding all valid (sub-)sentences for relation extraction.
The validity of a (sub-)sentence means the (sub-)sentence must cover the identified entity mentions.

An entity mention is defined as a span of tokens in a sentence.
The following sentence is an example in the NYT dataset. We refer to the sentence 
as $S1$: \\
$S1$: ``{\tt But that spasm of irritation by a master intimidator was minor compared with what
Bobby Fischer, the erratic former world chess champion, dished out in March at a news
conference in \underline{Reykjavik}, \underline{Iceland}.}''\\
The sentence is labeled with the \entity{`contains'} relation between two geographical locations.
The two underlined spans, {\tt \underline{Reykjavik}} and {\tt \underline{Iceland}}, 
are identified as the two entity mentions representing two locations.
 
There are simple sentences such as "{\tt \underline{Rechard Levine} was born in \underline{Manhattan}.}"  
It is labeled with the relation \entity{`place\_of\_birth'} between the two underlined entity mentions. 
There are also compound-complex
sentences consisting of multiple independent and dependent clauses. The NYT data set contains 372,853 
sentences. Out of them, 111,610 sentences are labeled with 24 relations defined in FreeBase \cite{freebase}.
For the labeled sentences, the longest sentence has 265 words, while the shortest sentence has 4 words. 
The mean length is 39 words. 
Since we will extract \emph{valid} sub-sentences to compare to the original sentences, 
we aimed to make sure that the original sentences are statistically long enough such 
that the extracted \emph{valid} 
sub-sentences are significantly different from the original ones.
To this extent, we extracted
sub-sentences consisting of the sequence of tokens between the two entity mentions
(including the two mentions.)
We call such 
a sub-sentence \emph{span}. 
Examining the lengths of spans, we find the longest span has 
99 words, while the shortest span has 2 words. The mean length of spans is 11 words. 
The standard deviation of the sentence lengths is 15 words while the
standard deviation of the span lengths is 9 words.
The dataset allows us to conduct the study on examining the embedding spaces of sentences and \emph{valid}
sub-sentences with significantly different lengths.

\begin{table*}[!th]
	\begin{center}
		\begin{tabular}{|l | l|}
			\hline
			\textbf{Method Name} & \textbf{Definition} \\
			\hline
			$X_{0}$:\textbf{span}: & \emph{extracts the sub-sentence starting from the first mention and ending at the second mention.}\\
			\hline
			$X_{1}$:\textbf{spanBA1}: & \emph{extracts the sub-sentence extending 1 token before the \textbf{span} and 1 word after the \textbf{span}.}\\
			$X_{2}$:\textbf{spanBA2}: & \emph{extracts the sub-sentence extending 2 tokens before the \textbf{span} and 2 words after the \textbf{span}.}\\
			... & ...  ...\\
			$X_{10}$:\textbf{spanBA10}: & \emph{extracts the sub-sentence extending 10 tokens before the \textbf{span} and 10 words after the \textbf{span}.}\\ 
			$X_{15}$:\textbf{spanBA15}: & \emph{extracts the sub-sentence extending 15 tokens before the \textbf{span} and 15 word after the \textbf{span}.}\\
			$X_{20}$:\textbf{spanBA20}: & \emph{extracts the sub-sentence extending 20 tokens before the \textbf{span} and 20 words after the \textbf{span}.}\\
			\hline 
			$Y_{1}$:\textbf{surroundings1}: & \emph{extracts the concatenation of the sub-sentences each of which containing 1 token}
			 \emph{surrounding a mention.}\\
			$Y_{2}$:\textbf{surroundings2}: & \emph{extracts the concatenation of the sub-sentences each of which containing 2 tokens} 
			\emph{surrounding a mention.}\\
			... & ... ...\\
			$Y_{10}$:\textbf{surroundings10}: & \emph{extracts the concatenation of the sub-sentences each of which containing 10 tokens} 
			\emph{surrounding a mention.}\\
			$Y_{15}$:\textbf{surroundings15}: & \emph{extracts the concatenation of the sub-sentences each of which containing 15 tokens} 
			\emph{surrounding a mention.}\\
			$Y_{20}$:\textbf{surroundings20}: & \emph{extracts the concatenation of the sub-sentences each of which containing 20 tokens} 
			\emph{surrounding a mention.}\\
			\hline
		\end{tabular}
	\end{center}
	\caption{\label{tab:sub-sentence-extraction} Sub-Sentence Extraction Methods}
\end{table*}
\comment{
Intuitively, a longer sentence describes multiple meanings corresponding to multiple semantic relations.
A similar issue called polysemy already exists in words. It has always been unclear how to interpret 
the embedding when the word in question is polysemous, that is, has multiple senses \cite{arora-polysemy}.
When a long sentence is encoded as a vector, it is even harder to make sense of it. The issue of 
polysemous words in NLP relies upon WordNet, a hand-constructed repository of word senses and their interrelationships. 
Unfortunately, there is no hope to develop such a repository for sentences. 
For the problem of applying embeddings for relation extraction, an idea is to
encode only necessary parts of a sentence. However, it is unclear what constitutes ``necessary parts'', 
and how the embedding methods will interact with sub-sentences. In this study, we take a first step 
to initially answer the questions.}

TABLE \ref{tab:sub-sentence-extraction}
summarizes the extraction methods.
Since all valid sub-sentences must contain the
identified entity mentions, the extractions anchor on the entity mentions. We also apply 
extraction of sub-sentences up to length 40, based on the average NYT sentence length of 39 tokens. 
Starting with the entity mentions, 
the first method,  $X_{0}$:\textbf{span}, extracts the sub-sentence spanning from the first entity mention to the 
second entity mention. The second method, $X_{1}$:\textbf{spanBA1}, extracts the sub-sentence extending the span
with one token before and after the entity mentions. Likewise, $X_{i}$:\textbf{spanBA$i$}, for $i=2...20$, extend
the span with $i$ tokens before and after the entity mentions. For instance, from the example sentence $S1$, 
$X_{0}$:\textbf{span} extracts ``{\tt \underline{Reykjavik}, \underline{Iceland}}",  $X_{1}$:\textbf{spanBA1} extracts
``{\tt in \underline{Reykjavik}, \underline{Iceland} .}", $X_{2}$:\textbf{spanBA2} extracts
``{\tt conference in \underline{Reykjavik}, \underline{Iceland} .}", and so on.
The next group of methods, $Y_{j}$:\textbf{surroundings$j$}, for $j=1...20$, extract the entity mentions
and $j$ tokens surrounding the entity mentions. 
The method $Y_{j}$:\textbf{surroundings$j$} differs from $X_{i}$:\textbf{spanBA$i$} in that
$Y_{j}$:\textbf{surroundings$j$} starts from entity mentions and extends on both sides of each mention, while
$X_{i}$:\textbf{spanBA$i$} starts from a span and extends on both sides of the span.
The method $Y_{j}$:\textbf{surroundings$j$} may extract discontinuous chunks from a sentence if
the two mentions are located far way from each other in the sentence, and  
will concatenate the discontinuous chunks as a single sub-sentence.
Finally, the methods extract all valid sub-sentences up to length around 40, though the sub-sentences may 
have duplicates extracted from short original sentences.  

%
%
\section{Analytic Design and Experimental Process}
\label{sec:analytic-design}

There are 9 embedding methods, and 42 sets of original sentences, spans, spanBA1-20, and surroundings1-20.
We conduct clustering and network analyses on $9\times 42=378$ embedding spaces generated by 
the combinations of embedding and sub-sentence extraction methods.
 
\subsection{Clustering Analysis}
\label{sec:clustering-analysis}

Clustering analysis \cite{interpretability-refinement-clustering} 
evaluates the extent to which the sentences expressing the same relations are located in the same partitions.
The analysis encompasses two main tasks \cite{data-mining-textbook}: 
(1) \emph{clustering tendency}
assesses whether it is suitable to apply clustering on the embedding spaces in first place, 
and (2) \emph{clustering evaluation} seeks to assess the 
goodness or quality of the clustering given data labels. 

\textbf{Clustering Tendency:} The metric we implemented for clustering tendency is 
\emph{Spatial Histogram (SpatHist)} \cite{data-mining-textbook,cluster-theory, clusterability}. 
Given a dataset $D$ with $d$ dimensions, we create $b$ equi-width bins
along each dimension, and count how many points lie in each of the $b^{d}$ $d$-dimensional cell.
We can obtain the empirical joint probability mass function (EPMF) of $D$ based on 
the binned data. Next, we generate $t$ random samples, each comprising $n$ uniformly generated points within the same 
$d$-dimensional space as the input data $D$. We can compute the EPMF for each sample too. Finally, we can measure how much 
the EPMF of the input data $D$ differs from the EPMF of each random sample
using the Kullback-Leibler (KL) divergence which is non-negative.
The KL divergence is zero when the input data behaves the same as the random sample. 
The SpatHist is the average of $t$ KL divergences between $D$ and the $t$ random samples.
The larger the SpatHist is, the more clusterable the data should be. 

\textbf{Clustering Evaluation.}  
Given a dataset $D$ with partitions $P=\{P_{1}, ..., P_{m}\}$ each of which has a label, a metric of clustering evaluation 
measures the extent to which points from the same partition appear 
in the same cluster, and the extent to which points from different partitions are grouped in different clusters.
The higher the metric value is, the better the quality of the clustering. 
For example, \emph{homogeneity} \cite{clustering-homogeneity} 
quantifies the extent to which a cluster contains entities from only
one partition. Suppose the data $D$ is clustered into 
$k$ groups $C=\{C_{1}, C_{2}, ..., C_{k}\}$. Let $N_{ij}$ be the number of 
members in group $C_{i}$ with partition label $j$. The homogeneity of the clustering is defined in terms of entropy
as: $h=1-\frac{H(C|P)}{H(P)}$, where $H(C|P)$ is the 
conditional entropy of the clustering $C$ given the partition
$P$ and $H(P)$ is the entropy of the original data partitioning. 
Other metrics apply the same principle but use different ways to measure cluster memberships.
We tested the following metrics: 
\emph{purity, fMeasure, Rand Index, homogeneity, mutual information, completeness, vMeasure,} and 
\emph{Fowlkes-Mallows measure}\footnote{https://scikit-learn.org/stable/modules/classes.html\#module-sklearn.metrics.cluster}.

For all the metrics, the higher the value is, the better the clustering quality. 
Our experiments show that these metrics are consistent
in terms of measuring the quality of clustering, though they may be at different value scales.

\subsection{Network Analysis}
\label{subsec:network-analysis}

For a set of sentence embeddings, we
build a Sentence Embedding Similarity Graph (SESG) based on the Euclidean
distances between embedding vectors. Given a set of sentences 
$\mathcal{S} =\{S_{1}, S_{2}, ..., S_{p}\}$, let
$\emb{W}=\{\emb{v}(S_{1}), \emb{v}(S_{2}), ..., \emb{v}(S_{p})\}$ be the 
set of sentence embeddings. We build the SESG graph corresponding to
the sentence embeddings as follows. Let $G=(V, E)$ be the SESG grpah, where
$V=\{v_{1}, v_{2}, ...., v_{n}\}$ is the set of vertices, and 
$E=\{e_{1}, e_{2}, ..., e_{m}\}$ is the set of edges.
Initially, the graph $G$ is empty. 
For a pair of embeddings $\emb{v}(S_{i}), \emb{v}(S_{j})\in \emb{W}$, 
we add two vertices $v_{i}, v_{j}\in V$ and an edge $e_{k}=(v_{i}, v_{j})\in E$ between them, 
if the distance between the embeddings of corresponding sentences is smaller than 
a threshold, i.e., $||\emb{v}(S_{i})-\emb{v}(S_{j})|| < thresh$.  In this study, we choose
the threshold as the \emph{mean Euclidean distance} between the embedding vectors of the sentences
that are labeled with the same relations. It should be noted that 
by cutting off pairs of sentences with
larger distances, not every sentence will have a corresponding 
vertex in the SESG graph.   

\subsection{Experimental Process}
\label{sec:experimental-process}

Each embedding spaces has 111,160 embedding vectors. We ran the experiments in multiple local and remote compute
instances, including 2 local machines each with 16G RAM, 2 virtual machines each with 32G RAM and 8 vCPU in an on-premises 
cloud, and a Google Colab Pro account. We ran the network analyses using Apache Spark 
on a Databricks cluster with 8 worker nodes of Amazon m4.large instance.  

The sentences in the NYT dataset are labeled by 24 Freebase relations. 
The sentences labeled with the same relation are considered in the same cluster. 
To recover the 24 clusters, 
we apply the K-Means implementation in Scikit-Learn package with $n\_clusters=24$ and 
other options with the default values. For clustering 
tendency, we use 500 random samples to average KL divergences for the final 
SpatHist values. The dimensions
of embeddings generated by the 9 embedding methods range from 300 to 4096. 
The main limitation of the spatial histogram is when we bin 
each dimension to create cells for computing the EPMF, the number of cells is exponentially large and 
most of the cells will be empty. To mitigate the problem, we apply a PCA ($n\_component = 2$) 
dimensionality reduction on 
embedding vectors before we compute the KL divergences.  

\begin{figure}[!th]
	\centering
	\includegraphics[width=1\linewidth]{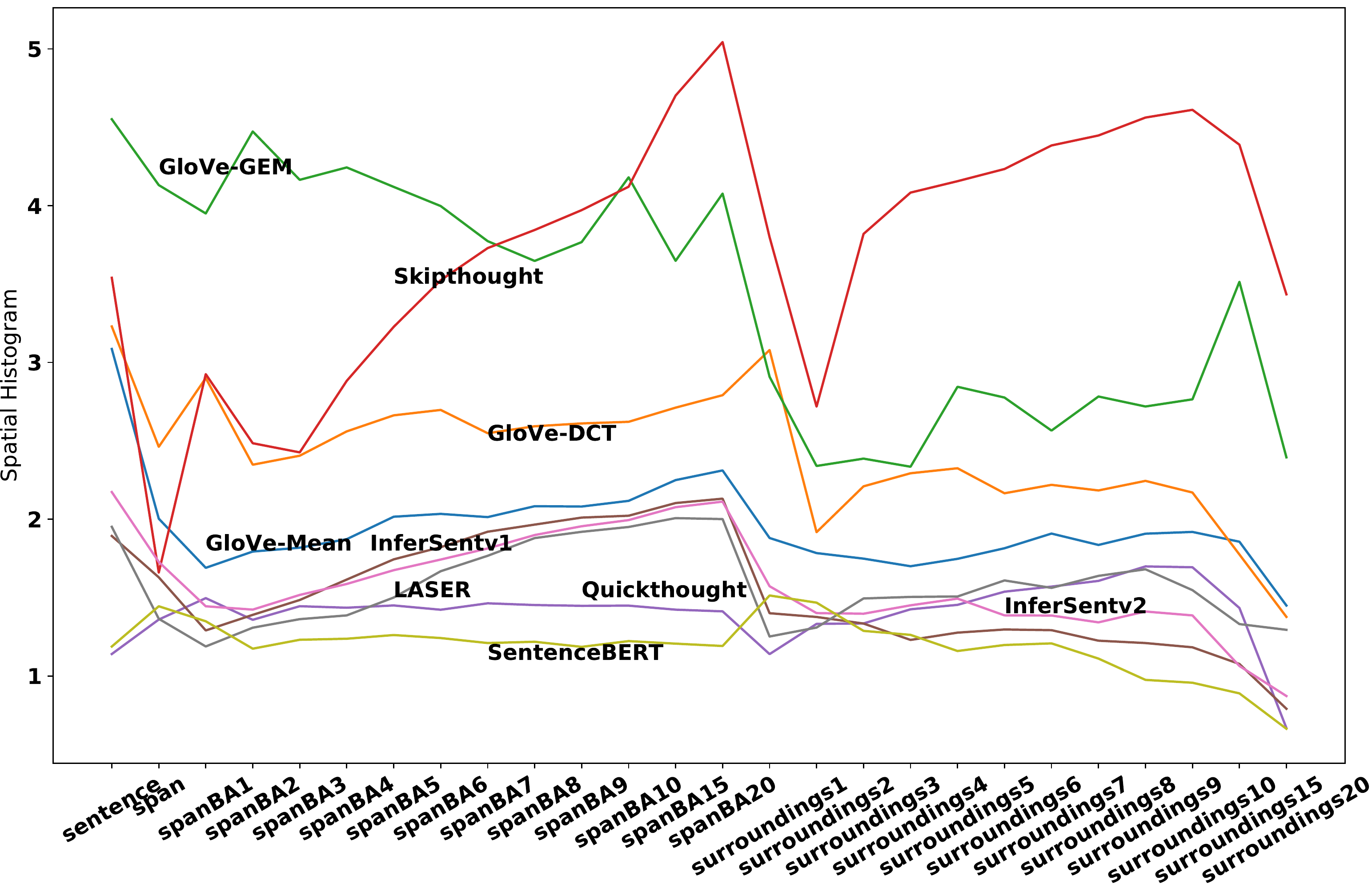} 
	\caption{Clustering tendency analysis: Each line represents an embedding method. 
		The x-axis lists the sets of sentences and sub-sentences used to generate the embeddings.
		The y-axis indicates the values of the metrics (spatial histogram) measuring the clusterability of the 
		embeddings. The higher the y values, the more clusterable the embeddings 
		corresponding to the sets of sentences and sub-sentences on the x-axis. 		
	}
	\label{fig:all-clustering-tendency}
\end{figure}
\begin{figure*}[!th]
	\centering
	\includegraphics[width=1\linewidth]{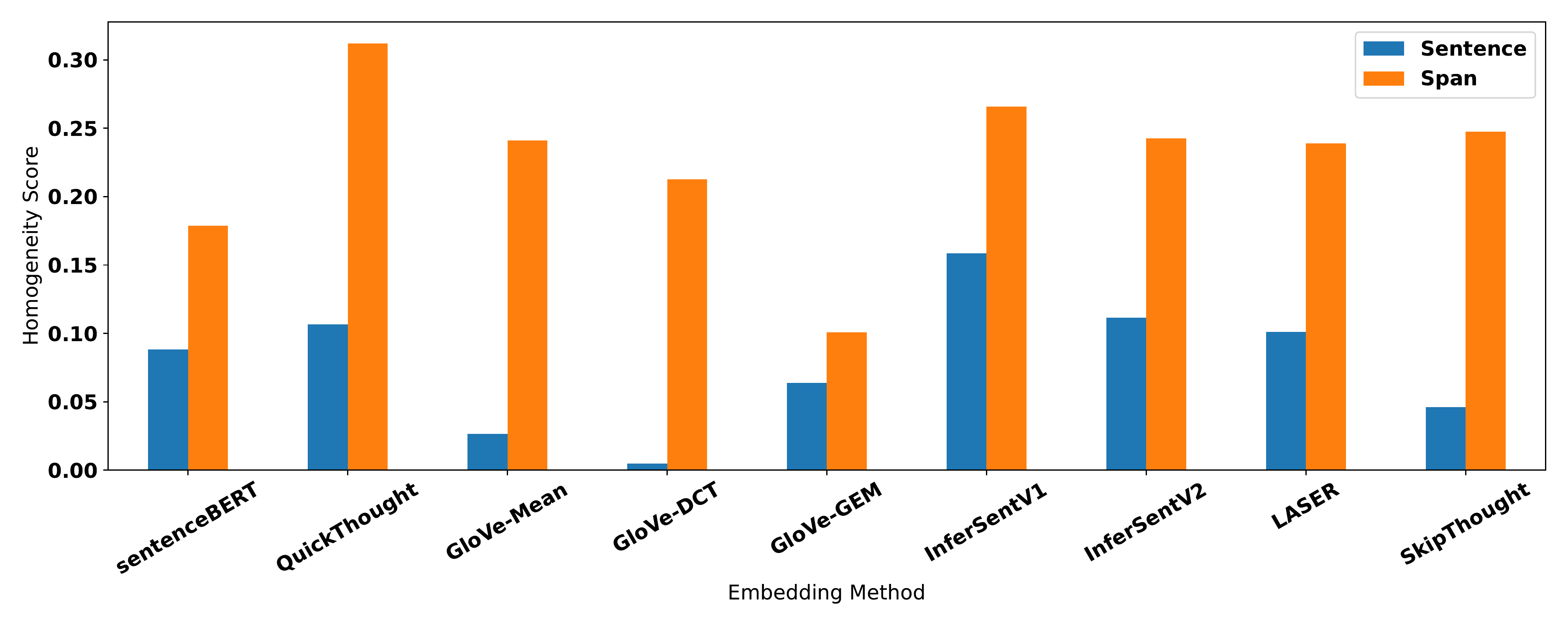}
	\caption{Clustering evaluation analysis:
	This figure illustrates the clustering evaluation results measured by the \emph{homogeneity score} metric. It shows the embeddings of the span sub-sentences are more clusterable than that of the original sentences. All metrics demonstrated the same property on different types of sub-sentences. The entire data are available in the github repository. 
    }
	\label{fig:clustering-validation-all}
\end{figure*}

%
%
\section{Experimental Results}
\label{sec:experimental-results}

\subsection{Clustering Tendency}

Figure \ref{fig:all-clustering-tendency} shows the results of clustering tendency analysis 
for the embedding spaces. Each value of the metric (spatial histogram) 
is computed by averaging 500 KL divergences. 
Their standard deviations are between $[0.005, 0.2]$ with a mean $0.02$. 
Here are some observations:
\begin{itemize}
	\item GloVe-GEM generates the most clusterable embeddings on the original sentences and spans with up to
	5 extra tokens. The clusterability of the embeddings generated by GloVe-GEM is better than most of the 
	other methods.
	\item Skipthought generates the most clusterable embeddings on the sub-sentences based on the tokens surrounding 
	entity mentions.
	\item SentenceBERT and Quickthought generate more clusterable embeddings on spans than on original sentences 
	(the lower-left corner area on the figure). 
\end{itemize}

\subsection{Clustering Evaluation}
Figure \ref{fig:clustering-validation-all} 
illustrates the clustering evaluation results measured by the \emph{homogeneity score} metric for
the embedding spaces of sentences and spans. It shows the embeddings of the span sub-sentences are more clusterable than that of the original sentences. All metrics demonstrated the same property on different types of sub-sentences. 
Looking into the entire data set, 
we have the following key observations:
\begin{itemize}
	\item Quickthought, GloVe-DCT, GloVe-GEM, and Skip-thought generate the embeddings with 
	better clustering quality on spans than on all other types of sentences. 
	\item SentenceBERT, InferSentV1, InferSentV2, and LASER generate the embeddings with
	better clustering quality on surroundings1 than on all other types of sentences.
	\item All of the embeddings generated by the methods from the original sentences
	are of low clustering quality. Some of them are with the worst clustering quality.
\end{itemize}
\begin{figure*}[!ht]
	\centering
	\subfloat[\entity{A Small Neighborhood of two hubs in GEM-Sentence}]{
		\includegraphics[width=0.5\linewidth]{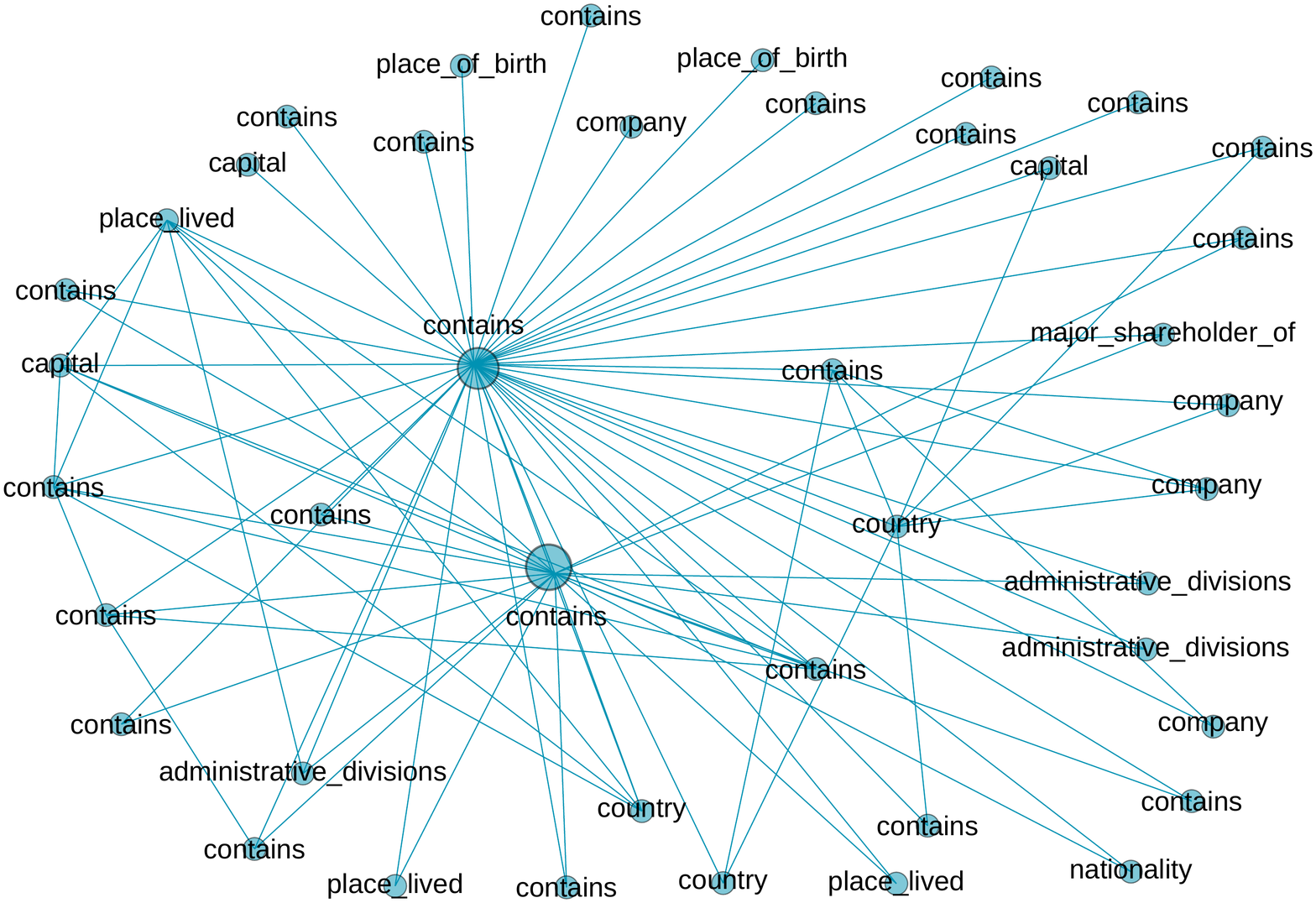} 
	}
	\subfloat[\entity{A Small Neighbordhood of two hubs in GEM-span}]{
		\includegraphics[width=0.5\linewidth]{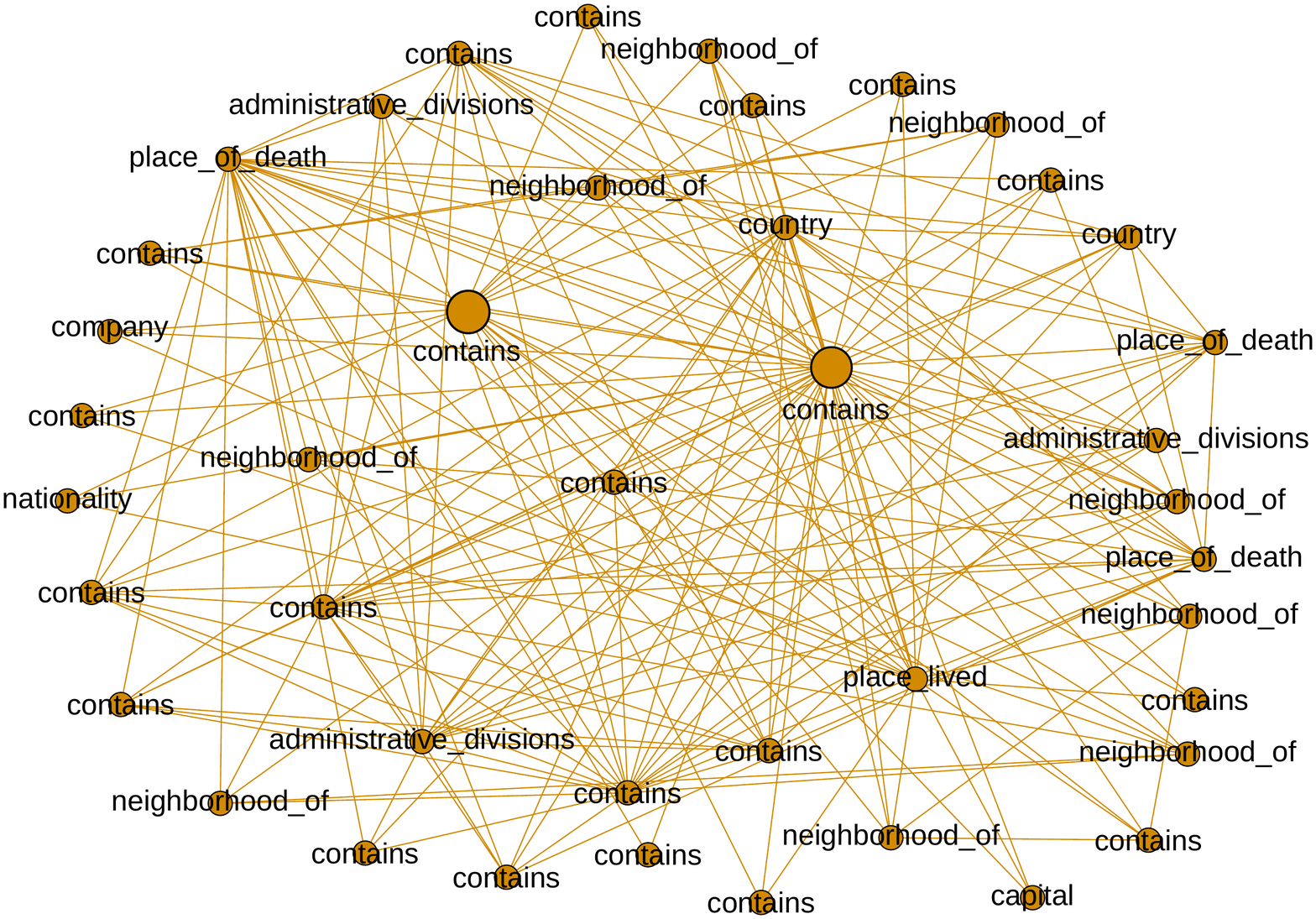}
	}
	\caption{Visualization of small neighborhood graphs. The vertices correspond to the embeddings
		of sentences in (a) and sub-sentences in (b). Each vertex is labeled by the relation which is used to label the sentences and 
		sub-sentences. 
		Each neighborhood graph is built around two hubs, 
		i.e., nodes with the highest degree. According to the construction of SESG graph, a hub is close to 
		its neighbors. If the neighbors are also close to each other, we should see dense connection
	in the neighborhood. The denser the neighborhood graph, the better the clusterability of the embeddings.}
	\label{fig:lcc-highest-degree-graph}
\end{figure*} 

\subsection{Network Analysis} 

The potential size of a Sentence Embedding Similarity Graph (SESG) is astronomically large. 
The number of  undirected pairs of sentences is 
about $111610^2/2 \approx 6.228\times10^{9}$.  It is infeasible to analyze the SESG graphs for 
all sets of (sub-)sentences. Because GloVe-GEM has the best performance shown by the clustering analyses, 
we choose the embeddings generated by GloVe-GEM on the original sentences and spans. We call these
two representative SESG graphs \entity{GEM-sentence} and \entity{GEM-span}, respectively.
Our primary goal is to enrich the findings of clustering analysis through a more
focused case study.
The final \entity{GEM-sentence} graph has 94,473 vertices and about $87$ million edges, 
and the final \entity{GEM-span} graph has 91,318 vertices and about $104$ million edges. The first observation is that the SESG graph built on 
span embeddings has more similar pairs than the graph built on sentence embeddings. We measures
the density of a graph by the ratio $\frac{number\_of\_edges}{number\_of\_total\_possible\_edges}$. The density
of \entity{GEM-span} is 25\% more than the density of \entity{GEM-sentence}.    

The first network analysis is on degree distributions. In both graphs, the degree distributions display 
a heavy tail. However, there are 543 vertices in \entity{GEM-span} having the highest degree (=31620), 
while there are only 25 vertices in \entity{GEM-sentence} having degrees greater than 31620. It indicates 
the similarity space of sentences is dominated by a few sentences. We randomly picked up two sentences with 
the highest degrees. They are ``{\tt Of Bronxvill, New York.}" and ``{\tt Of Plandome, New York.}" 
\emph{Both sentences closely mirror our definition of a span.}

The second analysis is about connected components. Both \entity{GEM-sentence} and \entity{GEM-span} have 
a large connected component (CC) with 60\% of vertices and more than 99\% of edges. About 40\% of their vertices
fall into other 12,400 in \entity{GEM-sentence} and 11,500 in \entity{GEM-span} 
smaller connected components. The density of the largest CC in \entity{GEM-span} is 35\% more than the 
density of the largest  CC in \entity{GEM-sentence}. \emph{Connected 
component density increases in comparison to the whole graph.}

The third analysis is aimed at the diameter of connected components. It is infeasible
to compute the shortest paths between all pairs of vertices in these big graphs ($\sim 10^9$). We randomly select
0.1\% of the vertices and compute the shortest paths. The longest distance is 8. 
However, in the random sets from both graphs, more than 87\% of the distances are shorter than 3. 
\emph{SESG graphs exhibit small world behavior.} 

The fourth analysis focuses on the relation distributions in the largest CC. 
There are 24 relations in the NYT dataset.
About 48\% of the sentences are labeled as the \entity{`contains'} relation between two geographical locations.
In the largest CC of \entity{GEM-sentence}, 50\% of the vertices corresponding to \entity{`contains'}. In the
largest CC of \entity{GEM-span}, 55\% of the vertices corresponding to \entity{'contains'}. \emph{This shows
that $X_{0}$:\textbf{span} improves the quality of the largest cluster of the embeddings.} 

Finally, we visualize small neighborhoods of the vertices with highest degree in 
Figure \ref{fig:lcc-highest-degree-graph}. From each graph, we randomly select
two vertices (the big dots) with the highest degree. For each selected vertex, we randomly
choose about 20 neighbors. The density of \emph{the small neighborhood graph from 
\entity{GEM-span} in Figure \ref{fig:lcc-highest-degree-graph} (b) is 16\%, while the density of  the small neighborhood graph from 
\entity{GEM-sentence} in Figure \ref{fig:lcc-highest-degree-graph} (a)} is 10\%. The former is denser than the latter one.  

%
%
\section{Discussion}
\label{sec:discussion}

We are motivated by the application of recognizing 
semantic relations between two entities in 
a textural sentence. Therefore, using the widely-used data set
for relation extraction research in the literature is well-suited for our purpose.
Given a set of embedding vectors as a metric space, the distance between any two members, which are usually
called points, characterizes the geometric structures of the space. 
Our analytical methods, specifically clustering and network analyses based 
on pair-wise distances, reveal a diversity of underlying geometric structures. 

Unlike word embedding methods, sentence embedding methods are guided by different underlying principles.
However, experimental results show the guiding principles may or may not converge on
generating embeddings with similar properties.
For example, the embedding spaces generated
by SentenceBERT and Quickthought on spans or short segments containing two entity mentions are more clusterable than
on the original sentences. It is interesting to note 
that SentenceBERT and Quickthought share little in terms of their modeling and training processes. 
It is important to note, however, that both Quickthought and Skipthought share the same principle that uses embeddings to predict the neighboring sentences. But the embedding spaces generated by them exhibit different geometric structures. The 
values of the clusterability metric in  
Figure \ref{fig:all-clustering-tendency} indicate that 
the methods GloVe-GEM, Skipthought, and GloVe-DCT are the candidates for generating embeddings with better clustering properties in real-world applications. 

%
%
\section{Conclusion}
\label{sec:conclusion}

In this study, we investigate the clusterability of embedding spaces generated by 
various sentence embedding methods on sentences and different sub-sentences. 
The primary motivation of the study is that more clusterable embeddings with 
better clustering quality capture more syntactic and semantic regularities. 
As a result, downstream NLP applications such as relation extraction 
would benefit from the embeddings with better clustering quality. 
We conduct a set of comprehensive clustering and network analyses on 
the embeddings generated by 9 main embedding methods.
The results show that the method GloVe-GEM stands out when applied to the original sentences and spans 
up to a certain length. Other methods have different strengths on different types of sub-sentences.
In most cases, the embeddings generated from the original sentences
are of low clustering quality. It signifies the impacts of 
sentence structures on the quality of embeddings when used by downstream applications.
The outcomes of our analysis can be used to aid and direct future sentence embedding models
and applications, for example,
combining the strengths of different embedding methods.   

\section*{Acknowledgment}

This work is supported in part by the USA NSF grant NSF-OAC 1940239.


\end{document}